\documentclass[letterpaper, 10 pt, conference]{ieeeconf/ieeeconf}  
\usepackage{graphicx}
\usepackage{afterpage}
\usepackage{amsmath}
\usepackage{amsfonts}
\usepackage{tikz}
\usetikzlibrary{arrows.meta, positioning, fit, calc}
\usepackage{hyperref}
\usepackage{soul}
\usepackage{booktabs}
\usepackage{censor}
\usepackage{comment}
\usepackage[caption=false,font=footnotesize]{subfig}
\usepackage[normalem]{ulem}

\newif\ifanonymize
\anonymizefalse 

\ifanonymize
  \usepackage{censor}
  \let\anon\censor
\else
  \newcommand{\anon}[1]{#1}
\fi

\usepackage{packages/herofigure}
\usepackage[font=small]{caption}

\hypersetup{
  colorlinks = true,
  linkcolor  = blue,
  urlcolor   = blue,
  citecolor  = blue
}

\IEEEoverridecommandlockouts                              

\title{\LARGE \bf
\texttt{SmallSatSim}: A GPU-Accelerated Microgravity Robotics Toolkit for Planning, Control, and Policy Learning
}

\author{\anon{David Schwartz$^{1}$, Alexander Hansson$^{1}$, Sabrina Bodmer$^{1}$,
David Sternberg$^{2}$,\\Oliver Jia-Richards$^{3}$, and Keenan Albee$^{4}$}%
\thanks{\anon{\raggedright$^{1}$ETH Z\"{u}rich, Rämistrasse 101, 8092 Zürich, Switzerland;} {\tt \anon{david.schwartz@alumni.ethz.ch;} \anon{alexander.hansson@mrl.ethz.ch; sabodmer@ethz.ch}}}%
\thanks{\anon{$^{2}$Jet Propulsion Laboratory, California Institute of} \anon{Technology, 4800 Oak Grove Drive, Pasadena, CA 91011, USA;} {\tt \anon{david.c.sternberg@jpl.nasa.gov}}}%
\thanks{\anon{$^{3}$University of Michigan, 500 South State Street, Ann Arbor, MI 48109,} \anon{USA; {\tt oliverjr@umich.edu}}}%
\thanks{\anon{$^{4}$University of Southern California, 3551 Trousdale Parkway, Los} \anon{Angeles, CA 90089, USA; {\tt kalbee@usc.edu}}}%
}

\begin{document}

\maketitle
\thispagestyle{empty}
\pagestyle{empty}

\begin{abstract}
Microgravity rendezvous and close proximity operations (RPO) is a growing area of interest for applications spanning in-space assembly and manufacturing (ISAM), orbital debris remediation, and small body exploration. Developing autonomy for these operations requires integrating dynamics simulation with task definition, control and learning algorithms, robustness testing, and evaluation. We present \texttt{SmallSatSim}, an open-source toolkit for developing autonomy algorithms for robots operating in microgravity environments.
Built on MuJoCo \cite{todorov2012mujoco}, \texttt{SmallSatSim} provides a common experiment and task abstraction for spacecraft models, planners, controllers and policies, actuator effects, disturbances, and evaluation, allowing model-based and learning-based approaches to operate on the same problem definitions. The framework combines conventional MuJoCo execution for model-based control with vectorized JAX/MJX execution for massively parallel policy learning.
We demonstrate \texttt{SmallSatSim} through Monte Carlo experiments on model-based control under actuator perturbations, five-seed PPO and SAC training under nominal and randomized dynamics with out-of-distribution evaluation, contact-rich rendezvous and docking, and GPU scaling experiments.\footnote{\texttt{SmallSatSim} can be found at \url{smallsatsim.github.io}.}
\end{abstract}

\section{Introduction}

Microgravity rendezvous and close proximity operations (RPO) allow robotic space systems to interact with other objects on-orbit in increasingly agile ways \cite{papadopoulosRoboticManipulationCapture2021} \cite{flores-abadReviewSpaceRobotics2014}. Proposed use cases have grown in recent years, including rendezvous with and inspection of space stations such as the Lunar Gateway \cite{fuller2022gateway, albeeArchitectingAutonomySafe2025}, on-orbit servicing and assembly \cite{maAdvancesSpaceRobots2023}, debris removal \cite{colvinCostBenefitAnalysis,arshad2025,Fallahiarezoodar2025}, and small body exploration \cite{nesnasAutonomousExplorationSmall2021}. These scenarios often involve complex dynamics, uncertainties, and the need to validate novel control strategies. Recent work has further emphasized robust and coordinated control for proximity operations \cite{Gong2024RobustPR}, collaborative servicing under partial feedback \cite{nino2025collaborative}, and learning-based approaches for agile rendezvous trajectory optimization and mission design \cite{guffanti2024transformers,takubo2025agile}. We focus on free-flyer dynamics in microgravity, as encountered during these close-range inspection and station-proximal operations. Realistic simulation environments are essential for developing and testing algorithms for these applications, as they allow researchers to evaluate performance under realistic conditions without the significant risk and cost associated with microgravity testing.

Realistic dynamics simulation alone, however, is not sufficient for developing autonomy for these scenarios. Researchers must formulate tasks and success criteria, instantiate spacecraft and actuation models, develop and interchange planners, controllers, and learned policies, introduce uncertainty and actuator failures, and evaluate different approaches under common conditions. Learning-based approaches additionally require scalable simulation across many randomized environments.

This work introduces \texttt{SmallSatSim}, a GPU-accelerated simulation toolkit for developing autonomy algorithms for robotic systems operating in microgravity environments. \texttt{SmallSatSim} builds on the MuJoCo \cite{todorov2012mujoco} physics engine. Tasks can be defined once and used with model-based controllers or learned policies. The toolkit includes spacecraft and actuation models, planning and control methods, failure and disturbance models, and tools for training and evaluation.

\texttt{SmallSatSim}'s primary contributions are:
\begin{enumerate}
    \item A common interface for defining spacecraft, tasks, planners, controllers, learned policies, and evaluation criteria.
    \item A JAX/MJX backend for batched spacecraft simulation and policy learning, supporting per-environment dynamics and actuator parameters and JIT-compiled rollouts.
    \item Tools for training and evaluation under uncertainty, including actuator failures, disturbances, domain randomization, logging, visualization, and checkpointing.
\end{enumerate}

Section \ref{sec:related_work} reviews related simulation environments and physics engines, including those specifically intended for RPO. Section \ref{sec:sim_env} details the architecture and features of \texttt{SmallSatSim}. Section \ref{sec:use_cases} evaluates \texttt{SmallSatSim} through model-based robustness, policy learning and domain randomization, contact-rich docking, and GPU-parallel simulation experiments. Finally, Section \ref{sec:conclusion} concludes with future directions and applications.

\section{Related Work}
\label{sec:related_work}

\subsection{Spacecraft and Orbital Simulation Frameworks}

Existing RPO robotics simulation frameworks include NASA's Astrobee \cite{smith2016astrobee} Gazebo simulation environment \cite{Fluckiger}, DLR's OOS-Sim \cite{artigasOOSSIMOngroundSimulation2015}, and high-fidelity orbital environment simulators like Tudat \cite{gisolfi2025open}, Ansys STK and GMAT \cite{hughes2017using}. Orsula et al. \cite{orsulaSpaceRoboticsBench2025} provide the most detailed overview to date of space robotic simulation environments, including RPO and microgravity simulators such as Basilisk \cite{kenneally2020basilisk}, as well as modular navigation frameworks such as Roboran \cite{el-hariry2025roboran}. More recently, mjorbit \cite{zhang2026mjorbit} combines orbital propagation with MuJoCo-based multibody and contact simulation and provides both CPU and GPU execution backends. Orbital and mission-level simulators generally emphasize long-horizon dynamics, reference frames, and environment modeling; robotics simulators instead focus on robot dynamics and task-level interaction.

\subsection{Space Robotics and Robot Learning Environments}

Two of the most widely used simulation stacks for robot dynamics learning and control are MuJoCo XLA (MJX), an official JAX/XLA-based accelerator backend for MuJoCo \cite{todorov2012mujoco}, and NVIDIA Isaac Lab \cite{mittal2025isaac}, a robot learning framework built on Isaac Sim \cite{NVIDIA_Isaac_Sim} and PhysX. MuJoCo
Playground \cite{zakka2025mujoco} builds on MuJoCo to provide GPU-accelerated environments and training tools for robot learning, while mjlab \cite{zakka2026mjlab} combines Isaac Lab-style environment abstractions with MuJoCo Warp. MJX-JAX supports automatic differentiation and JIT compilation through JAX, while Isaac Lab with PhysX provides a broader sensor and rendering stack.
More recently, Isaac Lab has added support for Newton \cite{santhosh2026accelerate}, a GPU-accelerated physics engine built on NVIDIA Warp. Space Robotics Bench (SRB) \cite{orsulaSpaceRoboticsBench2025} builds on the Isaac ecosystem to provide space-robotics environments with GPU-parallel simulation, domain randomization, sensor simulation and interfaces to established reinforcement learning libraries.

\begin{table}[t]
\centering
\vspace{2mm}
\caption{
Comparison of selected simulation stacks for robot learning.
\textbullet\ denotes native/full support,
$\circ$ partial or solver-dependent support,
and -- no native support.
}
\label{tab:physics_backends}

\renewcommand{\arraystretch}{1.18}
\setlength{\tabcolsep}{4pt}

\resizebox{\columnwidth}{!}{%
\begin{tabular}{
    @{}
    l
    c
    c
    c
    @{}
}
\toprule
\textbf{Systems property}
& \shortstack{\textbf{SmallSatSim} \\ \textbf{MuJoCo/MJX}}
& \shortstack{\textbf{Isaac Lab} \\ \textbf{Newton}}
& \shortstack{\textbf{Isaac Lab} \\ \textbf{PhysX}}
\\
\midrule

Massively parallel GPU simulation
& \textbullet
& \textbullet
& \textbullet
\\

JAX/XLA JIT-compilation
& \textbullet
& --
& --
\\

Automatic differentiation
& \textbullet
& $\circ$
& --
\\


\bottomrule
\end{tabular}%
}

\vspace{2pt}

\begin{minipage}{\columnwidth}
\footnotesize
\textit{Note:}
Newton supports differentiable physics, but differentiability depends on the selected solver. The primary validated Newton backend in Isaac Lab currently uses MuJoCo-Warp, which does not support automatic differentiation. The Newton backend in Isaac Lab is currently in beta.
\end{minipage}

\end{table}

\subsection{Positioning of \texttt{SmallSatSim}}

Table~\ref{tab:framework_comparison} compares \texttt{SmallSatSim} with representative space-robotics frameworks across their physical scope, simulation capabilities, and support for model-based and learning-based autonomy. \texttt{SmallSatSim} focuses on local microgravity autonomy development; long-horizon orbit and environment simulation and broad perception and sensor simulation are outside its current scope.

\begin{table*}[h]
\centering
\vspace{2mm}
\caption{
Comparison of simulation and autonomy-system capabilities across representative
space-robotics frameworks.
\textbullet\ denotes a reusable framework capability,
$\circ$ denotes a demonstrated example that is not provided as a reusable
framework component, and -- denotes a capability not provided or evidenced in
the public framework.
Per-environment randomization means that each parallel environment can use independently randomized spacecraft and actuator parameters.
}
\label{tab:framework_comparison}

\renewcommand{\arraystretch}{1.15}
\setlength{\tabcolsep}{6pt}

\begin{tabular}{
    @{}
    l
    c
    c
    c
    c
    c
    @{}
}
\toprule
\textbf{Capability}
& \textbf{SmallSatSim}
& \textbf{SRB}
& \textbf{mjorbit}
& \shortstack{\textbf{Basilisk} \\ \textbf{/ BSK-RL}}
& \shortstack{\textbf{Astrobee} \\ \textbf{Gazebo/FSW}}
\\
\midrule


\multicolumn{6}{@{}l}{\textit{Physical scope}}\\[-1pt]

\quad Local microgravity simulation
& \textbullet
& \textbullet
& \textbullet
& \textbullet
& \textbullet
\\

\quad Orbit/environment simulation
& --
& --
& \textbullet
& \textbullet
& --
\\

\quad Sensor simulation
& --
& \textbullet
& \textbullet
& \textbullet
& \textbullet
\\

\addlinespace[2pt]


\multicolumn{6}{@{}l}{\textit{Simulation systems}}\\[-1pt]

\quad Accelerator-parallel simulation
& \textbullet
& \textbullet
& \textbullet
& --
& --
\\

\quad Batched per-environment parameter randomization
& \textbullet
& \textbullet
& $\circ$
& --
& --
\\

\addlinespace[2pt]


\multicolumn{6}{@{}l}{\textit{Autonomy systems}}\\[-1pt]

\quad Reusable model-based autonomy
& \textbullet
& --
& $\circ$
& \textbullet
& \textbullet
\\

\quad Integrated policy-learning workflow
& \textbullet
& \textbullet
& $\circ$
& \textbullet
& --
\\

\bottomrule
\end{tabular}

\vspace{3pt}

\begin{minipage}{0.97\textwidth}
\footnotesize
\textit{Notes:}
Model-based autonomy denotes planning or control algorithms provided by the framework. BSK-RL is considered alongside Basilisk; SRB's policy-learning support uses integrated external RL libraries.
\end{minipage}

\end{table*}

\section{\texttt{SmallSatSim} Framework}
\label{sec:sim_env}

\subsection{Experiment and Task Abstraction}

Experiments specify a spacecraft model, task, planner or reference, controller or learned policy, uncertainty, and evaluation criteria. In Fig.~\ref{fig:sim_overview}, $\mathbf{x}_t$, $\mathbf{a}_t$,
$\mathbf{o}_t$, and $r_t$ denote the system state, action, observation, and
reward at time $t$, respectively, and $\boldsymbol{\theta}$ denotes the
environment parameters. Tasks define the observations, rewards, termination conditions, and success criteria used by different autonomy methods. The modular design allows new spacecraft, tasks, planners, controllers and policies to be added without modifying the simulator core.

\begin{figure*}[h]
\centering
\vspace{4pt}
\begin{tikzpicture}[
    scale=0.95,
    every node/.style={transform shape},
    font=\footnotesize,
    >=Latex,
    node distance=6mm,
    box/.style={
        draw,
        rounded corners,
        align=center,
        inner sep=4pt,
        minimum width=3.1cm
    },
    widebox/.style={
        box,
        minimum width=5.0cm
    },
    titlebox/.style={
        box,
        fill=black!3
    },
    line/.style={
        -{Latex},
        thick,
        shorten >=1pt,
        shorten <=1pt
    },
    dashedline/.style={
        -{Latex},
        thick,
        dashed,
        shorten >=1pt,
        shorten <=1pt
    }
]


\node[titlebox, minimum width=5.4cm] (user) {
    \textbf{Experiment definition}\\
    spacecraft, task, autonomy method,\\
    uncertainty, evaluation criteria
};


\node[widebox, below=17mm of user] (env) {
    \textbf{Task / environment API}\\[2pt]
    $\mathbf{x}_{t+1}
=f_{\boldsymbol{\theta}}(\mathbf{x}_t,\mathbf{a}_t)$\\
$\mathbf{o}_t
=h_{\boldsymbol{\theta}}(\mathbf{x}_t),\qquad
r_t=R_{\boldsymbol{\theta}}(\mathbf{x}_t,\mathbf{a}_t)$\\
    reset, termination, success logic
};


\node[box, left=19mm of env] (planner) {
    \textbf{Planner / reference interface}\\
    waypoint, inspection,\\
    custom planners
};

\node[box, right=19mm of env] (controller) {
    \textbf{Controller / policy interface}\\
    PD, LQR, MPC, GP-MPC,\\
    PPO, SAC, custom
};

\draw[line]
    ([yshift=3mm]planner.east) --
    node[above]{\small reference}
    ([yshift=3mm]env.west);

\draw[line]
    ([yshift=-3mm]env.west) --
    node[below]{\small state}
    ([yshift=-3mm]planner.east);

\draw[line]
    ([yshift=3mm]env.east) --
    node[above]{\small observations}
    ([yshift=3mm]controller.west);

\draw[line]
    ([yshift=-3mm]controller.west) --
    node[below]{\small action}
    ([yshift=-3mm]env.east);


\node[box, below left=12mm and 17mm of env] (effects) {
    \textbf{Failures, disturbances \& randomization}\\
    actuator maps $g_i(u)$, disturbances,\\
    randomized dynamics / parameters
};

\draw[dashedline]
    (effects.north east) --
    node[left, xshift=-2pt]{\small modify}
    (env.south west);


\node[
    draw,
    rounded corners,
    dashed,
    inner xsep=7pt,
    inner ysep=9pt,
    fit=(planner) (env) (controller) (effects),
    label={
        [font=\small, anchor=south west]
        north west:\textbf{\texttt{SmallSatSim} experiment API}
    }
] (apibox) {};

\draw[line]
    (user.south) --
    node[right]{\small configure}
    (apibox.north -| user.south);


\coordinate (apiCenterBelow)
    at ([yshift=-9mm]apibox.south);

\node[
    box,
    anchor=north,
    xshift=-22mm
] (cpu) at (apiCenterBelow) {
    \textbf{MuJoCo / CPU}\\
    conventional execution
};

\node[
    box,
    anchor=north,
    xshift=22mm
] (gpu) at (apiCenterBelow) {
    \textbf{MJX / JAX / GPU}\\
    vectorized execution
};

\coordinate (backendSplit)
    at ([yshift=5mm]apiCenterBelow);

\draw[thick]
    (env.south) -- (env.south |- backendSplit);

\draw[thick]
    (env.south |- backendSplit) -- (backendSplit);

\draw[line]
    (backendSplit) -| (cpu.north);

\draw[line]
    (backendSplit) -| (gpu.north);


\node[
    widebox,
    anchor=north,
    minimum width=5.2cm
] (lifecycle)
    at ([yshift=-10mm]$(cpu.south)!0.5!(gpu.south)$) {
    \textbf{Training and evaluation}\\
    logging, checkpoints, metrics,\\
    visualization, video
};

\coordinate (lifecycleSplit)
    at ([yshift=5mm]lifecycle.north);

\draw[thick]
    (cpu.south) -- (cpu.south |- lifecycleSplit);

\draw[thick]
    (gpu.south) -- (gpu.south |- lifecycleSplit);

\draw[thick]
    (cpu.south |- lifecycleSplit)
    -- (gpu.south |- lifecycleSplit);

\draw[line]
    (lifecycleSplit) -- (lifecycle.north);

\end{tikzpicture}

\caption{
\texttt{SmallSatSim}'s modular experiment architecture. A common task interface connects spacecraft models, planners, controllers or policies, failures, disturbances and randomization, and CPU or GPU simulation backends.
}
\label{fig:sim_overview}
\end{figure*}

\subsection{Spacecraft and Local Microgravity Models}

\texttt{SmallSatSim} models spacecraft motion in a local inertial frame under microgravity assumptions, with configurable spacecraft mass, inertia, and actuation. The default 6-DoF model used is based on Newton-Euler rigid body dynamics. The system state is defined as
\begin{equation}
\mathbf{x} = 
\begin{bmatrix}
\mathbf{r}^I_{CoM} & \mathbf{q}^{IB} & \mathbf{v}^B_{CoM} & \boldsymbol{\omega}^B
\end{bmatrix}^\top,
\end{equation}
where $\mathbf{r}^I_{CoM}$ is the position of the center of mass in the inertial frame, $\mathbf{q}^{IB}$ is a unit quaternion representing attitude, $\mathbf{v}^B_{CoM}$ is the linear velocity in the body frame, and $\boldsymbol{\omega}^B$ is the angular velocity in the body frame.

The dynamics evolve as:
\begin{align}
\dot{\mathbf{r}}^I_{CoM} &= \mathbf{R}_{IB} \mathbf{v}^B_{CoM}, \label{eq:pos} \\
\dot{\mathbf{q}}^{IB} &= \frac{1}{2} \mathbf{H}(\mathbf{q}^{IB})^\top \boldsymbol{\omega}^B, \label{eq:quat} \\
\dot{\mathbf{v}}^B_{CoM} &= \frac{1}{m} \mathbf{F}_{\text{thr}}^B - \boldsymbol{\omega}^B \times \mathbf{v}^B_{CoM}, \label{eq:linvel} \\
\dot{\boldsymbol{\omega}}^B &= \mathbf{I}^{-1} \left( \mathbf{T}_{\text{thr}}^B - \boldsymbol{\omega}^B \times (\mathbf{I} \boldsymbol{\omega}^B) \right), \label{eq:angvel}
\end{align}
with mass $m$, inertia matrix $\mathbf{I}$, and quaternion kinematic matrix $\mathbf{H}(\cdot)$. The force and torque produced by the thrusters are derived from:
\begin{equation}
\begin{bmatrix}
\mathbf{F}_{\text{thr}}^B \\
\mathbf{T}_{\text{thr}}^B
\end{bmatrix}
= \mathbf{B} \mathbf{u},
\end{equation}
where $\mathbf{u} \in \mathbb{R}^{n_u}$ contains the individual
thruster inputs and $\mathbf{B}\in\mathbb{R}^{6\times n_u}$ is the
thruster mixer matrix that maps the individual thrust commands to the net body-frame force and torque applied to the spacecraft. For thruster $i$ with body-frame thrust direction
$\mathbf{d}_i^B$ and position $\mathbf{r}_i^B$ relative to the center of
mass, the corresponding column of $\mathbf{B}$ is
\begin{equation}
    \mathbf{B}_{:,i}
    =
    \begin{bmatrix}
        \mathbf{d}_i^B \\
        \mathbf{r}_i^B \times \mathbf{d}_i^B
    \end{bmatrix}.
\end{equation}

\subsection{Planning and Control}

\texttt{SmallSatSim} provides common interfaces for planners and controllers. The toolkit includes proportional-derivative (PD), linear-quadratic regulator (LQR), model predictive control (MPC), model predictive contouring control (MPCC), Gaussian process-based MPC (GP-MPC) controllers, and RL policies, as well as waypoint and inspection planners. The planner, controller, and learning interfaces can also be extended with custom implementations.

\subsection{Failures, Disturbances, and Randomization}

To simulate actuator degradation and failure, \texttt{SmallSatSim} allows the injection
of nonlinear perturbations on a per-thruster basis. These perturbations modify
the demanded thrust $\mathbf{u}_{\text{dem}}$ into an actual applied thrust
$\mathbf{u}_{\text{act}}$ according to
\begin{equation}
u_{\text{act},i} = g_i(u_{\text{dem},i}),
\end{equation}
where $g_i(\cdot)$ is a possibly nonlinear and stochastic function representing
a perturbation model for thruster $i$. In the nominal case, $g_i(u) = u$.
These mappings are used to compute the actual force and torque applied to the
system via
\begin{equation}
\begin{bmatrix}
\mathbf{F}_{\text{thr}}^B \\
\mathbf{T}_{\text{thr}}^B
\end{bmatrix}
= \mathbf{B} \, \mathbf{u}_{\text{act}}.
\end{equation}

\texttt{SmallSatSim} supports several classes of $g_i(\cdot)$, including:
\begin{itemize}
    \item \textbf{Stuck-off}: $g_i(u) = 0$
    \item \textbf{Stuck-on}: $g_i(u) = u_{\max}$
    \item \textbf{Saturation}: $g_i(u) = \min(u, u_{\text{sat}})$
    \item \textbf{Faulty valve}: piecewise-linear approximation
    \item \textbf{Thrust instability}: sinusoidal + noise perturbation
    \item \textbf{GP-based nonlinear samples}: stochastic mappings sampled from a GP posterior
\end{itemize}

Disturbances can also be added directly to the dynamics to represent external forces and effects not captured by the nominal model.
In vectorized simulations, dynamics and actuator parameters
can be randomized independently for each environment.

\subsection{Contact and Docking}

\texttt{SmallSatSim} uses MuJoCo's collision and compliant contact models
\cite{todorov2012mujoco,todorovConvexAnalyticallyinvertibleDynamics2014} to simulate physical contact. Collision geometry
and contact parameters can be configured for different spacecraft and
environments. We demonstrate this capability in the contact-rich rendezvous
and docking experiment in Section~\ref{sec:use_cases}, where contact forces and
post-contact behavior are evaluated.

\subsection{Policy Learning and GPU Execution}

\texttt{SmallSatSim} provides two backends for RL environments: a lightweight
free-flyer dynamics backend and an MJX backend using the MuJoCo model. Both backends use the same vectorized environment interface. With MJX, the simulation step and rollout path can be just-in-time (JIT) compiled by JAX for massively parallel policy learning.

The toolkit provides native JAX implementations of Vanilla Policy Gradient (VPG) \cite{schulman2015high}, Proximal Policy Optimization (PPO) \cite{schulman2017proximal}, and Soft Actor-Critic (SAC) \cite{haarnoja2018soft}. Policies can be trained under nominal or randomized dynamics and evaluated on seeded nominal, in-distribution, out-of-distribution, and stress scenarios, as shown in Fig.~\ref{fig:rl}.

\begin{figure*}[h]
\centering
\vspace{4pt}

\resizebox{0.98\textwidth}{!}{%
\begin{tikzpicture}[
    font=\large,
    >=Latex,
    box/.style={
        draw,
        rounded corners,
        align=center,
        inner xsep=9pt,
        inner ysep=8pt,
        minimum height=1.45cm
    },
    head/.style={
        box,
        fill=black!3
    },
    line/.style={
        -{Latex},
        thick,
        shorten >=2pt,
        shorten <=2pt
    }
]


\node[
    box,
    minimum width=5.0cm
] (env) {
    \textbf{Batched randomized environments}\\[3pt]
    $\boldsymbol{\theta}^{(i)} \sim p_{\mathrm{train}}(\boldsymbol{\theta})$,
    \qquad $i=1,\ldots,N$\\
    mass/inertia $\cdot$ thrust $\cdot$ disturbances
};


\node[
    box,
    minimum width=4.4cm,
    right=11mm of env
] (learn) {
    \textbf{Accelerated policy learning}\\[3pt]
    PPO / SAC $\cdot$ nominal / randomized\\
    JAX/MJX $\cdot$ JIT rollouts $\cdot$ GPU
};


\node[
    box,
    minimum width=4.5cm,
    right=11mm of learn
] (eval) {
    \textbf{Common seeded evaluation}\\[3pt]
    Nominal: $\boldsymbol{\theta}=\boldsymbol{\theta}_0$\\
    ID: $\boldsymbol{\theta}\sim p_{\mathrm{train}}(\boldsymbol{\theta})$\\
    OOD: $\boldsymbol{\theta}\sim p_{\mathrm{OOD}}(\boldsymbol{\theta})$\\
    Stress: disturbance / compound shift
};


\node[
    head,
    minimum width=3.8cm,
    right=11mm of eval
] (metrics) {
    \textbf{Robustness metrics}\\[3pt]
    success rate\\
    $e_{\mathrm{pos}}$ $\cdot$ $e_{\mathrm{att}}$\\
    control effort
};


\draw[line]
    (env.east) --
    node[
        above,
        font=\normalsize,
        fill=white,
        inner sep=2pt
    ]{rollouts}
    (learn.west);

\draw[line]
    (learn.east) --
    node[
        above,
        font=\normalsize,
        fill=white,
        inner sep=2pt
    ]{freeze checkpoints}
    (eval.west);

\draw[line]
    (eval.east) --
    (metrics.west);


\coordinate (configCenter)
    at ($(env.north)!0.5!(learn.north)$);

\node[
    head,
    minimum width=8.5cm,
    minimum height=0.9cm,
    inner ysep=4pt,
    above=9mm of configCenter
] (config) {
    \textbf{RL experiment configuration}
    \qquad
    task $\cdot$ algorithm $\cdot$ regime
    $\cdot$ seeds $s=1,\ldots,5$
};

\coordinate (cfgsplit)
    at ([yshift=-4mm]config.south);

\draw[thick]
    (config.south) -- (cfgsplit);

\draw[line]
    (cfgsplit) -| (env.north);

\draw[line]
    (cfgsplit) -| (learn.north);

\end{tikzpicture}%
}

\caption{
Policy-learning and robustness-evaluation workflow used in our experiments.
Policies are trained under nominal or randomized dynamics and evaluated on common seeded nominal, ID, OOD, and stress scenarios.
}
\label{fig:rl}
\end{figure*}

\subsection{Evaluation, Logging, and Visualization}

Experiments can be evaluated over seeded scenarios with configurable metrics, and training runs can be checkpointed and resumed. Results can be logged using the native logger or Weights \& Biases, and simulations can be visualized using the MuJoCo viewer, browser-based viewer, or recorded to video.

\section{Experiments and Evaluation}
\label{sec:use_cases}

This section evaluates \texttt{SmallSatSim} through four experiments covering model-based control under actuator perturbations, policy learning under distribution shift, contact-rich docking, and GPU-parallel simulation. The
experiment configurations and scripts used for these evaluations are included
in the public codebase.

\subsection{Robust Model-Based Control under Actuator Perturbations}

\begin{figure}[!b]
  \centering
  \includegraphics[width=\linewidth]{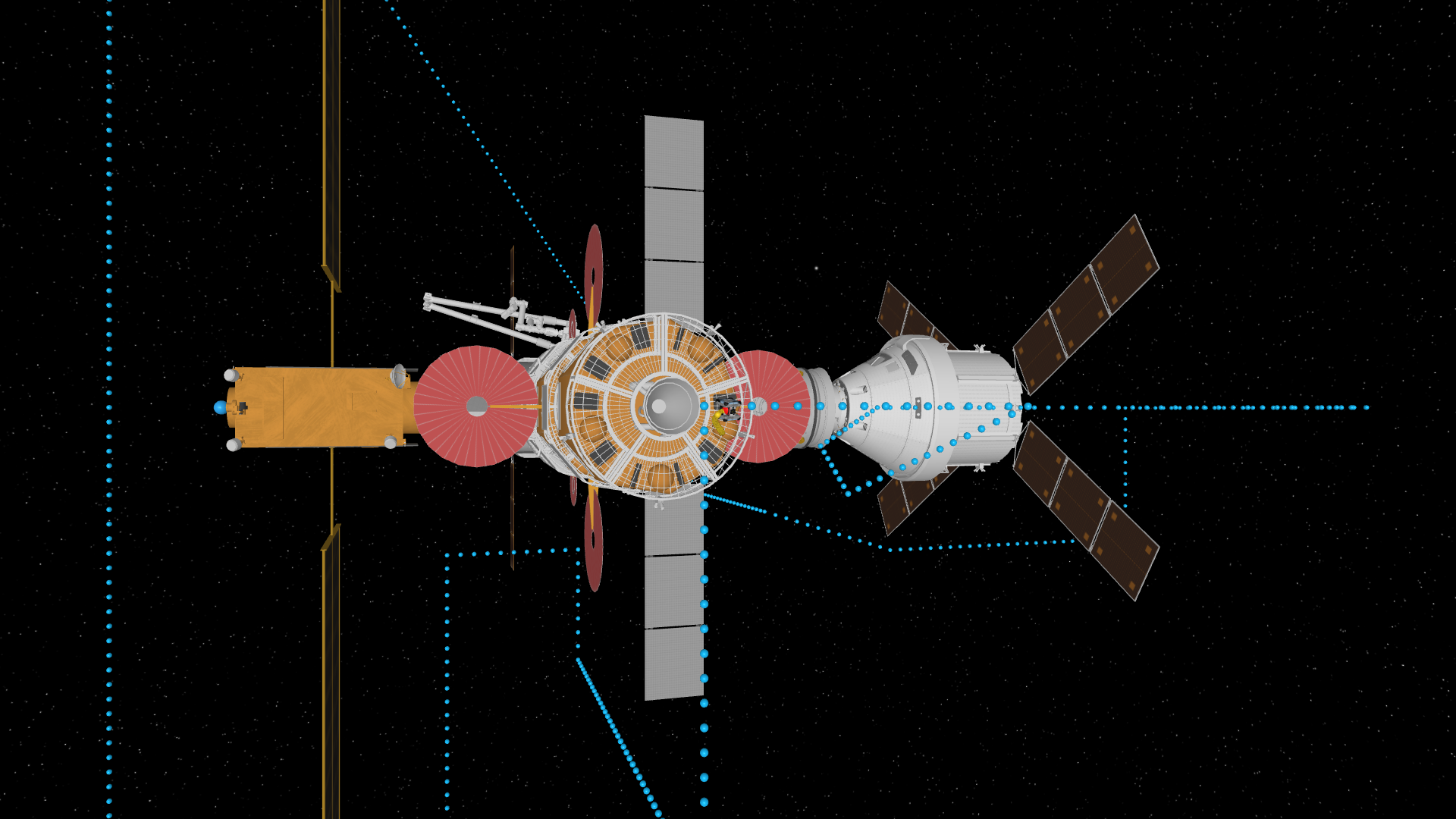}
  \caption{Lunar Gateway inspection by Astrobee, using the \texttt{NominalMPCController} and \texttt{MissionPlanner} inspection planner.}
  \label{fig:inspection} 
\end{figure}

As an example of model-based control in a close-proximity task,
\texttt{SmallSatSim} can be used for inspection of the Lunar Gateway with a 6-DoF free-flying robot operating near large-scale collision geometry (Fig.~\ref{fig:inspection}) \cite{albeeArchitectingAutonomySafe2025}. The task consists of station-keeping and waypoint-based inspection along reference trajectories produced by an inspection planner. The vehicle tracks full-pose setpoints in SE(3) using a nominal nonlinear MPC controller.

For systematic evaluation of robustness under actuator perturbations, we use a
controlled 6-DoF setpoint-regulation task with the same nonlinear MPC
controller. For each condition, 100 trials are run with randomized initial
states. We evaluate nominal operation, an external disturbance, faulty-valve
behavior, thrust saturation, stuck-off and stuck-on thrusters, and thrust
instability.

A trial is considered successful when the final position and attitude errors
are below $0.25$~m and $0.25$~rad, respectively, the linear and angular
velocities are below $0.15$~m/s and $0.05$~rad/s, respectively, and these
conditions are maintained for five consecutive steps.

\begin{figure}[!b]
  \centering
  \includegraphics[width=\linewidth]
  {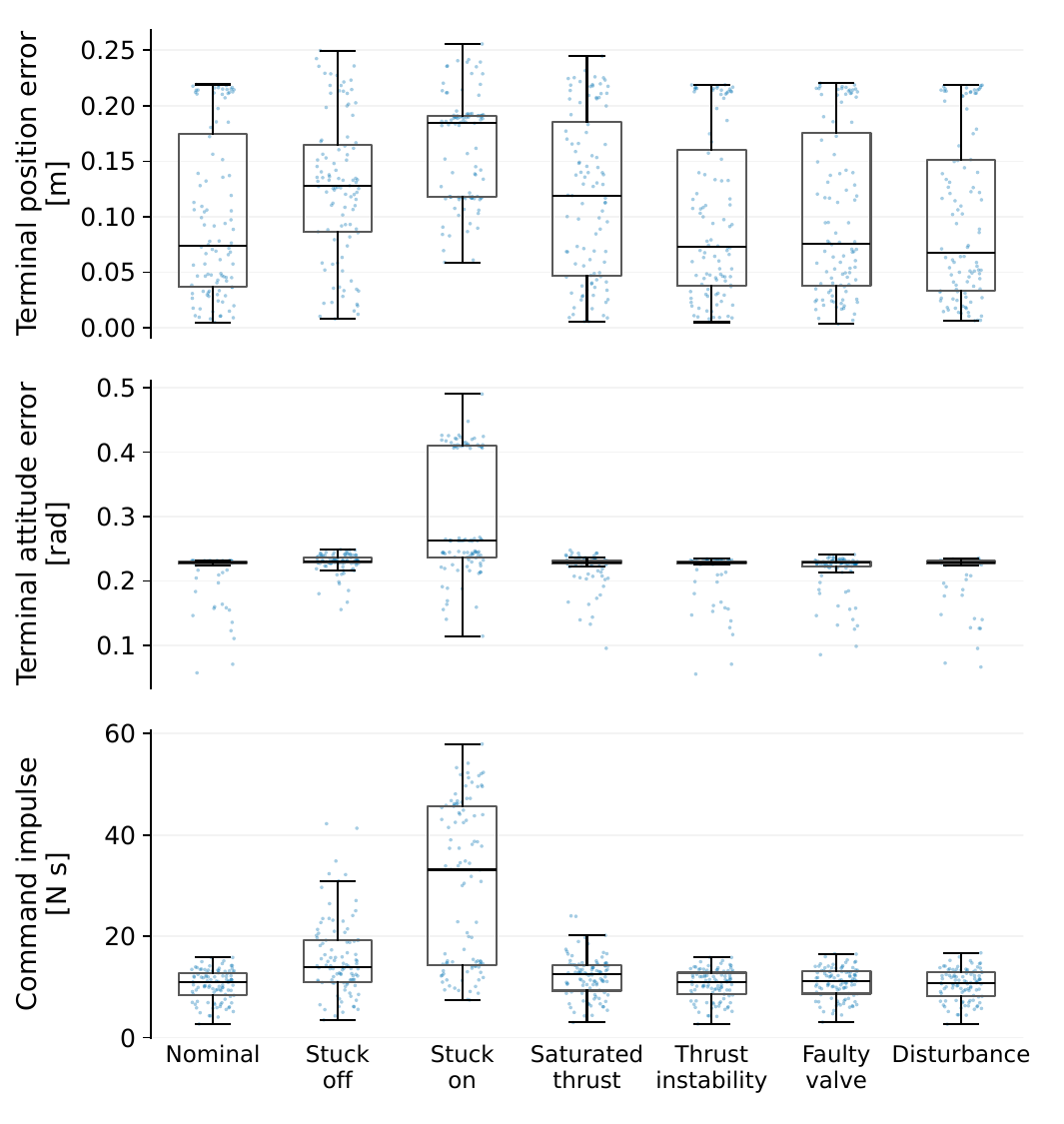}
  \caption{
  Monte Carlo robustness distributions for model-based control under actuator
  perturbations and disturbances.
  }
  \label{fig:model_based_robustness}
\end{figure}

\begin{table*}[h]
\centering
\caption{Model-based robustness over 100 trials per condition. Errors and
control effort are reported as median [Q1, Q3], where Q1 and Q3 denote the
25th and 75th percentiles.}
\label{tab:model_based_robustness}
\begin{tabular}{lcccc}
\toprule
\textbf{Condition}
& \textbf{Success}
& \textbf{Final pos. error [m]}
& \textbf{Final att. error [rad]}
& \textbf{Control effort [N s]} \\
\midrule
Nominal
& 100/100
& 0.074 [0.037, 0.174]
& 0.229 [0.228, 0.231]
& 10.97 [8.48, 12.82] \\

Disturbance
& 100/100
& 0.068 [0.033, 0.151]
& 0.229 [0.227, 0.231]
& 10.85 [8.28, 12.92] \\

Faulty valve
& 100/100
& 0.076 [0.038, 0.176]
& 0.228 [0.222, 0.230]
& 11.19 [8.70, 13.05] \\

Saturated thrust
& 100/100
& 0.119 [0.047, 0.185]
& 0.229 [0.227, 0.231]
& 12.62 [9.28, 14.29] \\

Stuck-off
& 100/100
& 0.128 [0.086, 0.164]
& 0.230 [0.228, 0.237]
& 14.00 [10.91, 19.23] \\

Stuck-on
& 45/100
& 0.185 [0.118, 0.191]
& 0.263 [0.237, 0.410]
& 33.15 [14.37, 45.70] \\

Thrust instability
& 100/100
& 0.073 [0.038, 0.160]
& 0.229 [0.227, 0.230]
& 11.03 [8.53, 12.84] \\
\bottomrule
\end{tabular}
\end{table*}

All conditions except stuck-on achieve a 100\% success rate; under stuck-on failures, this drops to 45\%. Stuck-off and thrust saturation
increase position error and control effort relative to nominal operation,
while stuck-on failures result in the largest degradation across the reported
metrics (Table~\ref{tab:model_based_robustness}).

\subsection{Policy Learning and Domain Randomization}

\begin{figure}[!t]
  \centering

  \includegraphics[width=\linewidth]
  {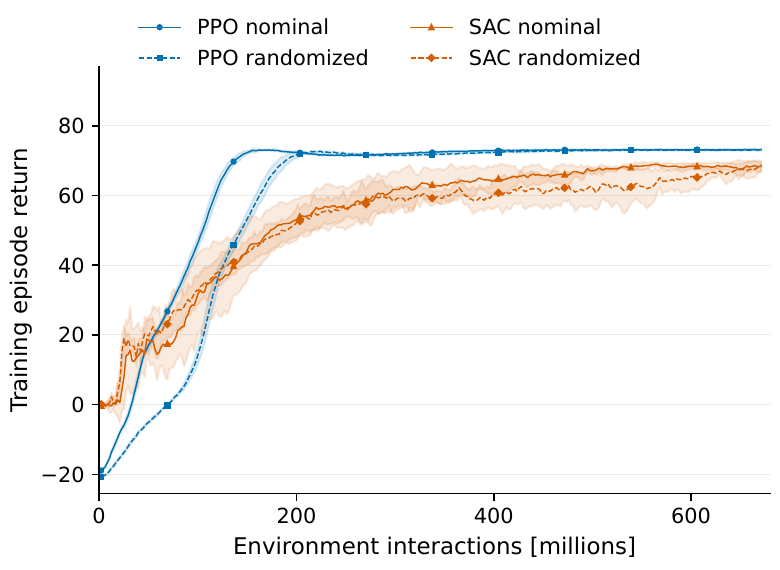}

  \vspace{2mm}

  \includegraphics[width=\linewidth]
  {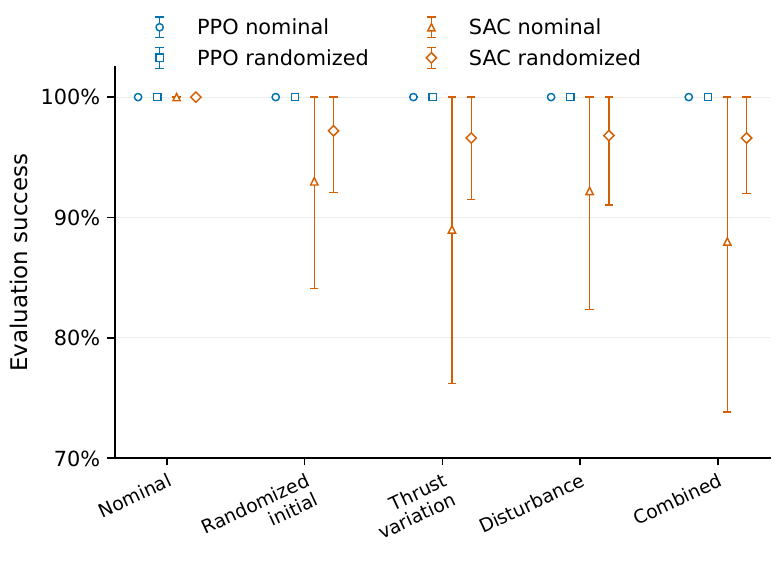}

  \caption{
  Policy learning across five training seeds and evaluation under nominal,
  randomized-initial-state, thrust-variation, disturbance, and combined
  conditions. Shaded bands show one standard deviation across training seeds;
  error bars show 95\% Student-$t$ confidence intervals across seeds.
  }
  \label{fig:rl_learning_shift}
\end{figure}

We compare PPO \cite{schulman2017proximal} and SAC \cite{haarnoja2018soft}
trained under nominal or randomized dynamics, using five training seeds per
configuration and the same interaction budget. Randomized training varies
thruster performance and external disturbances. Figure~\ref{fig:rl_learning_shift}
shows training episode return, averaged within 2.1-million-interaction windows
and across seeds, with shaded bands indicating one standard deviation. PPO
converges faster in both training regimes, while SAC improves more gradually.


After training, each policy is evaluated on a fixed suite consisting of a
deterministic nominal episode and 100 episodes each with randomized initial
states, thrust variation, external disturbances, and combined perturbations.
The same evaluation scenarios are used across training seeds. Both PPO
variants achieve 100\% observed success throughout the evaluation suite.
For SAC, randomized training improves mean success in every non-nominal
condition, with the largest difference under combined perturbations, where
success increases from 88.0\% to 96.6\%.

\subsection{Docking Contact Characterization}

\begin{figure}[!h]
  \centering
  \includegraphics[width=0.9\linewidth]{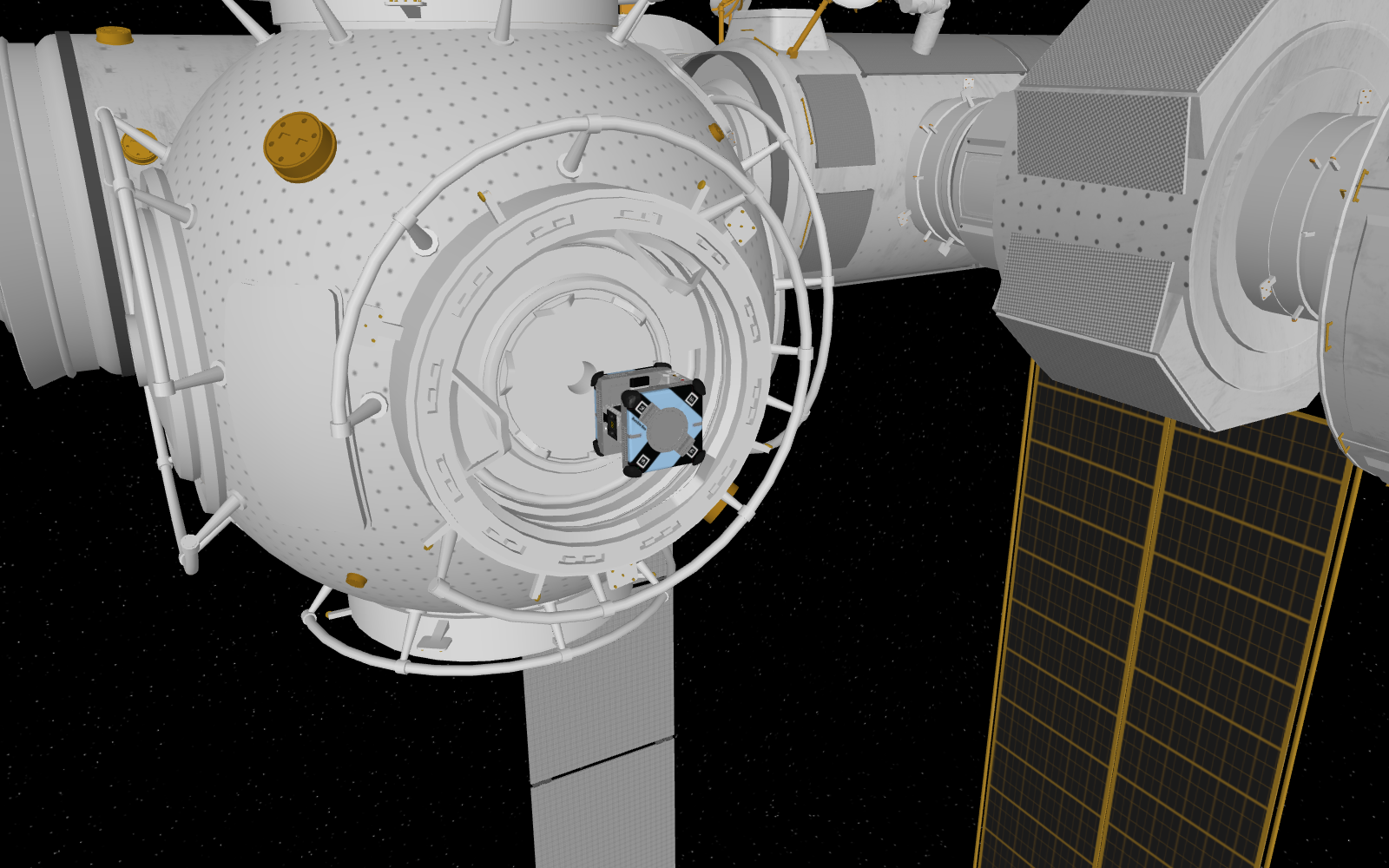}
  \caption{
  Rendezvous and docking with compliant contact at the Lunar Gateway.
  }
  \label{fig:docking_characterization}
\end{figure}

To demonstrate contact simulation, we consider rendezvous and docking at a
Lunar Gateway docking port. The vehicle is initialized with a randomized
relative pose and performs a staged approach consisting of pre-dock guidance
followed by final docking (Fig.~\ref{fig:docking_characterization}). The
simulation continues through physical contact using MuJoCo's compliant
contact model.

We evaluate 100 trials with randomized initial conditions under the default
contact parameters and additionally vary the contact parameters to assess
sensitivity. All 100 default trials result in a successful dock. In the
representative trial shown in Fig.~\ref{fig:docking_characterization}, first
contact occurs at $73.76$~s with a peak contact force of $5.73$~N. Three
contact events are recorded before the vehicle settles, illustrating that the
simulation captures the initial impact and subsequent re-contact rather than
terminating at first contact. Varying the contact parameters preserves docking
success over the tested settings while changing the measured peak contact
force.

\subsection{GPU-Parallel Simulation Performance}

For policy-learning experiments that do not require MuJoCo contact or scene
simulation, \texttt{SmallSatSim}'s lightweight free-flyer backend evaluates
the spacecraft dynamics directly in JAX while using the same vectorized
environment interface as the MJX backend. We benchmark this implementation
on an NVIDIA A100-SXM4-40GB GPU, varying the number of parallel environments
from 1 to 4096. The simulation timestep is $0.01$~s
with a control decimation of 5, such that each environment step advances
$0.05$~s of simulated time. Each measurement uses a rollout of 512 environment
steps and is repeated 10 times.

\begin{figure}[!t]
  \centering
  \includegraphics[width=\linewidth]
  {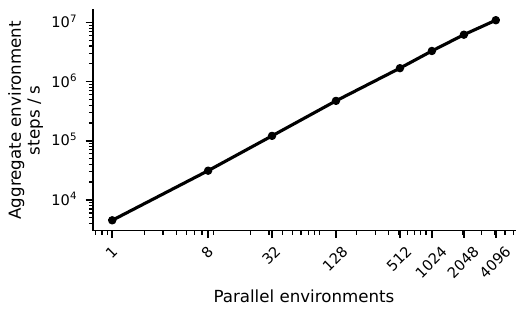}
  \caption{
  Aggregate rollout throughput with increasing numbers of parallel
  environments on an NVIDIA A100-SXM4-40GB GPU.
  }
  \label{fig:gpu_scaling_curve}
\end{figure}

\begin{table*}[!t]
\centering
\vspace{2mm}
\caption{
GPU rollout performance using the free-flyer dynamics backend on an NVIDIA
A100-SXM4-40GB GPU. Values are means over 10 repeats.
}
\label{tab:gpu_scaling}
\begin{tabular}{rcccc}
\toprule
\textbf{\# Envs}
& \textbf{Env Steps/s}
& \textbf{Aggregate Sim s/s}
& \textbf{Per-env Sim s/s}
& \textbf{Speedup} \\
\midrule
1
& $4.50\times10^{3}$
& 225.1
& 225.1
& $1.0\times$ \\

8
& $3.11\times10^{4}$
& 1556.0
& 194.5
& $6.9\times$ \\

32
& $1.21\times10^{5}$
& 6041.8
& 188.8
& $26.8\times$ \\

128
& $4.73\times10^{5}$
& 23635.1
& 184.6
& $105.0\times$ \\

512
& $1.69\times10^{6}$
& 84424.3
& 164.9
& $375.1\times$ \\

1024
& $3.30\times10^{6}$
& 164991.6
& 161.1
& $733.0\times$ \\

2048
& $6.22\times10^{6}$
& 310995.0
& 151.9
& $1381.6\times$ \\

4096
& $1.09\times10^{7}$
& 547112.2
& 133.6
& $2430.6\times$ \\
\bottomrule
\end{tabular}
\end{table*}

As shown in Table~\ref{tab:gpu_scaling}, aggregate throughput increases from approximately $4.50\times10^3$
environment steps/s with one environment to $1.09\times10^7$ environment
steps/s with 4096 environments. The corresponding per-environment simulation
rate decreases from $225.1$ to $133.6$ simulated seconds per wall-clock
second as the batch size increases. The lightweight backend is intended for large-scale policy learning and domain
randomization, while the MJX backend can be used when MuJoCo physics and contact simulation are required.

\section{Conclusion}
\label{sec:conclusion}

We have presented \texttt{SmallSatSim}, a simulation and training toolkit for developing and testing autonomy algorithms for microgravity robotic RPO. \texttt{SmallSatSim}
provides planning, control, and policy-learning methods together with tools for modeling failures and disturbances, evaluation, logging, and visualization, providing a significant upgrade in CPU- and GPU-compatible autonomy workflows for microgravity robotics.

The experiments demonstrate model-based control under a range of actuator
perturbations, policy learning and evaluation under randomized dynamics,
contact-rich rendezvous and docking, and parallel simulation across thousands
of environments. We believe \texttt{SmallSatSim} will facilitate simplified experimentation of a number of microgravity robotics autonomy scenarios of interest to the community, including on-orbit contact-rich interaction, online model learning under system failures, and more. In future work, we aim to expand \texttt{SmallSatSim}'s default environments to include manipulation, multi-agent coordination, and debris remediation tasks.

\section*{Acknowledgment}

Initial development was supported by \anon{a NASA University} \anon{Smallsat Technology Partnership (USTP) program award} \anon{80NSSC23M0237}. The authors would like to thank \anon{Dr. Amir Rahmani} for his group's early support of \texttt{SmallSatSim}.\\
The authors used OpenAI ChatGPT and Codex to assist with software development, including code generation, debugging, and refactoring, as well as figure preparation. All generated content was reviewed and validated by the authors.

\bibliographystyle{ieeeconf/IEEEtran}
\bibliography{references}

@inproceedings{todorov2012mujoco,
  title={MuJoCo: A physics engine for model-based control},
  author={Todorov, Emanuel and Erez, Tom and Tassa, Yuval},
  booktitle={2012 IEEE/RSJ International Conference on Intelligent Robots and Systems},
  pages={5026--5033},
  year={2012},
  organization={IEEE},
  doi={10.1109/IROS.2012.6386109}
}

@misc{NVIDIA_Isaac_Sim,
    author = {{NVIDIA}},
    license = {Apache-2.0},
    title = {{Isaac Sim}},
    url = {https://github.com/isaac-sim/IsaacSim},
    version = {5.0.0}
}

@article{schulman2017proximal,
  title={Proximal policy optimization algorithms},
  author={Schulman, John and Wolski, Filip and Dhariwal, Prafulla and Radford, Alec and Klimov, Oleg},
  journal={arXiv preprint arXiv:1707.06347},
  year={2017}
}

@article{schulman2015high,
  title={High-dimensional continuous control using generalized advantage estimation},
  author={Schulman, John and Moritz, Philipp and Levine, Sergey and Jordan, Michael and Abbeel, Pieter},
  journal={arXiv preprint arXiv:1506.02438},
  year={2015}
}

@article{el-hariry2025roboran,
title={Robo{RAN}: A Unified Robotics Framework for Reinforcement Learning-Based Autonomous Navigation},
author={Matteo El-Hariry and Antoine Richard and Ricard Marsal and Luis Felipe Wolf Batista and Matthieu Geist and C{\'e}dric Pradalier and Miguel Olivares-Mendez},
journal={Transactions on Machine Learning Research},
issn={2835-8856},
year={2025},
url={https://openreview.net/forum?id=0wDbhLeMj9},
note={}
}

@article{arshad2025,
author = {Muneeb Arshad and Michael C.F. Bazzocchi and Faraz Hussain},
journal = {Acta Astronautica},
pages = {996-1022},
title = {Emerging strategies in close proximity operations for space debris removal: A review},
volume = {228},
year = {2025}
}

@article{Gong2024RobustPR,
  title={Robust Proximity Rendezvous and Coordinated Control of Space Robots},
  author={Kai Gong},
  journal={Advances in Space Research},
  year={2024},
  url={https://api.semanticscholar.org/CorpusID:273805435}
}

@article{takubo2025agile,
  title={Agile Tradespace Exploration for Space Rendezvous Mission Design via Transformers},
  author={Takubo, Yuji and Gammelli, Daniele and Pavone, Marco and D'Amico, Simone},
  journal={arXiv preprint arXiv:2510.03544},
  year={2025}
}

@inproceedings{guffanti2024transformers,
  title={Transformers for trajectory optimization with application to spacecraft rendezvous},
  author={Guffanti, Tommaso and Gammelli, Daniele and D’Amico, Simone and Pavone, Marco},
  booktitle={2024 IEEE Aerospace Conference},
  pages={1--13},
  year={2024},
  organization={IEEE}
}

@article{nino2025collaborative,
  title={Collaborative Spacecraft Servicing Under Partial Feedback Using Lyapunov-Based Deep Neural Networks},
  author={Nino, Cristian F and Patil, Omkar Sudhir and Petersen, Christopher D and Phillips, Sean and Dixon, Warren E},
  journal={The Journal of the Astronautical Sciences},
  volume={72},
  number={4},
  pages={40},
  year={2025},
  publisher={Springer}
}

@article{Fallahiarezoodar2025,
author = {Nasim Fallahiarezoodar and Zheng H. Zhu},
journal = {Space: Science \& Technology},
pages = {0291},
title = {Review of Autonomous Space Robotic Manipulators for On-Orbit Servicing and Active Debris Removal},
volume = {5},
year = {2025}
}

@article{kenneally2020basilisk,
  title={Basilisk: A flexible, scalable and modular astrodynamics simulation framework},
  author={Kenneally, Patrick W and Piggott, Scott and Schaub, Hanspeter},
  journal={Journal of aerospace information systems},
  volume={17},
  number={9},
  pages={496--507},
  year={2020},
  publisher={American Institute of Aeronautics and Astronautics}
}

@article{gisolfi2025open,
  title={Open-Source High-Fidelity Orbit Estimation for Planetary Science and Space Situational Awareness Using the Tudat Software},
  author={Gisolfi, Luigi and Dirkx, Dominic and Fayolle, Sam and Filice, Valerio and Alkahal, Riva and Avillez, Miguel and Dijkstra, Tristan and Hener, Jonas and Hin{\"u}ber, Lars and Langbroek, Marco and others},
  journal={arXiv preprint arXiv:2510.23179},
  year={2025}
}

@inproceedings{smith2016astrobee,
  title={Astrobee: A new platform for free-flying robotics on the international space station},
  author={Smith, Trey and Barlow, Jonathan and Bualat, Maria and Fong, Terrence and Provencher, Christopher and Sanchez, Hugo and Smith, Ernest},
  booktitle={International Symposium on Artificial Intelligence, Robotics, and Automation in Space (i-SAIRAS)},
  number={ARC-E-DAA-TN31584},
  year={2016}
}

@article{fuller2022gateway,
  title={Gateway program status and overview},
  author={Fuller, Sean and Lehnhardt, Emma and Zaid, Christina and Halloran, Kate},
  journal={Journal of Space Safety Engineering},
  volume={9},
  number={4},
  pages={625--628},
  year={2022},
  publisher={Elsevier}
}

@inproceedings{todorovConvexAnalyticallyinvertibleDynamics2014,
  title = {Convex and Analytically-Invertible Dynamics with Contacts and Constraints: {{Theory}} and Implementation in {{MuJoCo}}},
  shorttitle = {Convex and Analytically-Invertible Dynamics with Contacts and Constraints},
  booktitle = {2014 {{IEEE International Conference}} on {{Robotics}} and {{Automation}} ({{ICRA}})},
  author = {Todorov, Emanuel},
  year = 2014,
  month = may,
  pages = {6054--6061},
  publisher = {IEEE},
  address = {Hong Kong, China},
  doi = {10.1109/ICRA.2014.6907751},
  urldate = {2025-11-11},
  isbn = {978-1-4799-3685-4},
  langid = {english}
}

@inproceedings{albeeArchitectingAutonomySafe2025,
  title = {Architecting {{Autonomy}} for {{Safe Microgravity Free-Flyer Inspection}}},
  booktitle = {{{IEEE Aerospace}}},
  author = {Albee, Keenan and Sternberg, David C and Hansson, Alexander and Schwartz, David and Majumdar, Ritwik and {Jia-Richards}, Oliver},
  year = 2025,
  copyright = {All rights reserved},
  langid = {english}
}

@article{artigasOOSSIMOngroundSimulation2015,
  title = {The {{OOS-SIM}}: {{An}} on-Ground Simulation Facility for on-Orbit Servicing Robotic Operations},
  author = {Artigas, Jordi and De Stefano, Marco and Rackl, Wolfgang and Lampariello, Roberto and Brunner, Bernhard and Bertleff, Wieland and Burger, Robert and Porges, Oliver and Giordano, Alessandro and Borst, Christoph and {Albu-Schaeffer}, Alin},
  year = 2015,
  journal = {Proceedings - IEEE International Conference on Robotics and Automation},
  volume = {2015-June},
  number = {June},
  pages = {2854--2860},
  publisher = {IEEE},
  issn = {10504729},
  doi = {10.1109/ICRA.2015.7139588},
  isbn = {9781479969234}
}

@misc{colvinCostBenefitAnalysis,
  title = {Cost and {{Benefit Analysis}} of {{Orbital Debris Remediation}}},
  author = {Colvin, Thomas J and Karcz, John and Wusk, Grace and Besha, Patrick and Naasz, Bo},
  langid = {english}
}

@article{flores-abadReviewSpaceRobotics2014,
  title = {A Review of Space Robotics Technologies for On-Orbit Servicing},
  author = {{Flores-Abad}, Angel and Ma, Ou and Pham, Khanh and Ulrich, Steve},
  year = 2014,
  journal = {Progress in Aerospace Sciences},
  volume = {68},
  pages = {1--26},
  issn = {03760421},
  doi = {10.1016/j.paerosci.2014.03.002},
  urldate = {2017-11-02},
  isbn = {0376-0421}
}

@inproceedings{Fluckiger,
  title = {Astrobee Robot Software: Enabling Mobile Autonomy on the {{ISS}}},
  booktitle = {Proc. of the {{Int}}. {{Symposium}} on {{Artificial Intelligence}}, {{Robotics}} and {{Automation}} in {{Space}} (i-{{SAIRAS}})},
  author = {Fl{\"u}ckiger, Lorenzo and Browne, Kathryn and Coltin, Brian and Fusco, Jesse and Morse, Theodore and Symington, Andrew},
  year = 2018,
  urldate = {2018-08-10}
}

@article{maAdvancesSpaceRobots2023,
  title = {Advances in {{Space Robots}} for {{On}}-{{Orbit Servicing}}: {{A Comprehensive Review}}},
  shorttitle = {Advances in {{Space Robots}} for {{On}}-{{Orbit Servicing}}},
  author = {Ma, Boyu and Jiang, Zainan and Liu, Yang and Xie, Zongwu},
  year = 2023,
  month = aug,
  journal = {Advanced Intelligent Systems},
  volume = {5},
  number = {8},
  pages = {2200397},
  issn = {2640-4567, 2640-4567},
  doi = {10.1002/aisy.202200397},
  urldate = {2025-11-09},
  langid = {english}
}

@article{nesnasAutonomousExplorationSmall2021,
  title = {Autonomous {{Exploration}} of {{Small Bodies Toward Greater Autonomy}} for {{Deep Space Missions}}},
  author = {Nesnas, Issa A. D. and Hockman, Benjamin J. and Bandopadhyay, Saptarshi and Morrell, Benjamin J. and Lubey, Daniel P. and Villa, Jacopo and Bayard, David S. and Osmundson, Alan and Jarvis, Benjamin and Bersani, Michele and Bhaskaran, Shyam},
  year = 2021,
  month = nov,
  journal = {Frontiers in Robotics and AI},
  volume = {8},
  pages = {650885},
  issn = {2296-9144},
  doi = {10.3389/frobt.2021.650885},
  urldate = {2025-11-09},
  langid = {english}
}

@misc{orsulaSpaceRoboticsBench2025,
  title = {Space {{Robotics Bench}}: {{Robot Learning Beyond Earth}}},
  shorttitle = {Space {{Robotics Bench}}},
  author = {Orsula, Andrej and Geist, Matthieu and {Olivares-Mendez}, Miguel and Martinez, Carol},
  year = 2025,
  month = sep,
  number = {arXiv:2509.23328},
  eprint = {2509.23328},
  primaryclass = {cs},
  publisher = {arXiv},
  doi = {10.48550/arXiv.2509.23328},
  urldate = {2025-11-09},
  archiveprefix = {arXiv},
  langid = {english}
}

@article{papadopoulosRoboticManipulationCapture2021,
  title = {Robotic Manipulation and Capture in Space: A Survey},
  shorttitle = {Robotic Manipulation and Capture in Space},
  author = {Papadopoulos, Evangelos and Aghili, Farhad and Ma, Ou and Lampariello, Roberto},
  year = 2021,
  month = jul,
  journal = {Frontiers in Robotics and AI},
  volume = {8},
  pages = {686723},
  issn = {2296-9144},
  doi = {10.3389/frobt.2021.686723},
  urldate = {2022-02-10},
  langid = {english}
}

@article{zhang2026mjorbit,
  title={mjorbit: A Simulation Framework for Space Robotics},
  author={Zhang, John Z and Verhagen, Joris and Vega, Fausto and McKeen, Patrick and Manchester, Zachary},
  journal={arXiv preprint arXiv:2609.08010},
  year={2026}
}

@inproceedings{hughes2017using,
  title={Using the general mission analysis tool (GMAT)},
  author={Hughes, Steven P and Conway, Darrel J and Parker, Joel},
  booktitle={Aas guidance and control conference},
  number={GSFC-E-DAA-TN39043},
  year={2017}
}

@article{zakka2026mjlab,
  title={mjlab: A lightweight framework for gpu-accelerated robot learning},
  author={Zakka, Kevin and Liao, Qiayuan and Yi, Brent and Lay, Louis Le and Sreenath, Koushil and Abbeel, Pieter},
  journal={arXiv preprint arXiv:2601.22074},
  year={2026}
}

@article{zakka2025mujoco,
  title={MuJoCo playground},
  author={Zakka, Kevin and Tabanpour, Baruch and Liao, Qiayuan and Haiderbhai, Mustafa and Holt, Samuel and Luo, Jing Yuan and Allshire, Arthur and Frey, Erik and Sreenath, Koushil and Kahrs, Lueder A and others},
  journal={arXiv preprint arXiv:2502.08844},
  year={2025}
}

@article{mittal2025isaac,
  title={Isaac lab: A gpu-accelerated simulation framework for multi-modal robot learning},
  author={Mittal, Mayank and Roth, Pascal and Tigue, James and Richard, Antoine and Zhang, Octi and Du, Peter and Serrano-Munoz, Antonio and Yao, Xinjie and Zurbr{\"u}gg, Ren{\'e} and Rudin, Nikita and others},
  journal={arXiv preprint arXiv:2511.04831},
  year={2025}
}

@inproceedings{santhosh2026accelerate,
  title={Accelerate Robot Learning With NVIDIA Isaac Lab and Newton},
  author={Santhosh, Akul and Chadha, Rishabh},
  booktitle={Proceedings of the Special Interest Group on Computer Graphics and Interactive Techniques Conference Courses},
  pages={1--2},
  year={2026}
}

@article{haarnoja2018soft,
  title={Soft actor-critic algorithms and applications},
  author={Haarnoja, Tuomas and Zhou, Aurick and Hartikainen, Kristian and Tucker, George and Ha, Sehoon and Tan, Jie and Kumar, Vikash and Zhu, Henry and Gupta, Abhishek and Abbeel, Pieter and others},
  journal={arXiv preprint arXiv:1812.05905},
  year={2018}
}

\end{document}